\documentclass{article}
\usepackage{spconf,amsmath,graphicx}
\usepackage{amsmath, amssymb, mathrsfs}
\usepackage{subcaption}

\title{An Hybrid Quantum-Classical Diffusion Model for Image Generation}

\name{Qipeng Qian\textsuperscript{1}, Keli Deng\textsuperscript{2}, Yuntao Qian\textsuperscript{2}\iffalse\thanks{This work was supported in part by the National Key Research and Development Program of China under Grant 2023YFE0204200 and in part by the National Natural Science Foundation of China under Grant 62071421.}\fi}
\address{
\textsuperscript{1}Program of Applied Mathematics, Department of Mathematics, University of Arizona, Tucson, USA\\
\textsuperscript{2}College of Computer Science and Technology, Zhejiang University, Hangzhou, China
}
\begin{document}
%
\maketitle
\begin{abstract}
Quantum diffusion models provide a physics-consistent route to generative learning by formulating noising and denoising directly on quantum states.
However, applying such models to classical high-dimensional data is constrained by the qubit cost of state encoding and the computational burden of simulating large density operators.
We propose a scalable hybrid generative pipeline that combines a classical autoencoder for dimensionality reduction with a mixed-state quantum denoising diffusion probabilistic model (MSQuDDPM) operating in the learned latent space.
The autoencoder compresses data into compact latent codes that can be embedded into a small-qubit Hilbert space, after which the quantum diffusion model learns a generative distribution over latent density operators and decodes samples back to the original domain.
Algorithmically, we simplify the reverse dynamics by predicting an estimate of the clean state $\rho_0$ at timestep $t$ and computing the one-step reverse update via an analytic backward propagation rule, rather than learning an explicit predictor for $\rho_{t-1}$.
We demonstrate the proposed approach on MNIST image generation and discuss how mixed-state quantum diffusion can serve as a practical backbone for hybrid quantum--classical generative modeling under realistic qubit budgets.
\end{abstract}


\section{Introduction}
\label{sec:intro}

Denoising diffusion probabilistic models (DDPMs) have become a central paradigm for modern generative modeling by learning a Markovian reverse process that progressively transforms simple noise into structured samples \cite{ho2020denoising,song2019generative}. This formulation has proved attractive in practice because it is stable to train and can achieve high sample quality, but it is also computationally demanding: sampling typically requires evaluating a denoiser across many timesteps, and the cost grows quickly with the dimensionality of the sample space. 

Motivated by the search for more efficient generative mechanisms, quantum machine learning (QML) leverages variational quantum circuits and hybrid quantum--classical training, with recent work extending quantum models to structured data such as circuit-implementable quantum graph learning architectures (e.g., QSGCN \cite{zheng2024quantum} and quantum empowered GNNs for hyperspectral change detection \cite{lin2024quantum}). On the generative side, QGANs \cite{lloyd2018quantum,dallaire2018quantum} and hybrid variants have been explored for imaging tasks including hyperspectral restoration \cite{lin2025hyperking} and higher-resolution synthesis \cite{ma2025quantum}, while diffusion-style quantum generators remain relatively scarce and largely confined to small-scale benchmarks \cite{kolle2024quantum}. A complementary direction therefore formulates diffusion directly over quantum states and channels, where forward noising and reverse denoising are physically valid CPTP maps and the reverse model preserves positivity and unit trace; QuDDPM \cite{zhang2024generative} and its mixed-state extension MSQuDDPM \cite{kwun2025mixed} exemplify this approach by enabling generative learning over quantum ensembles while accounting for mixedness from decoherence, partial information, or stochastic preparation.

However, using quantum diffusion for classical generation faces a key bottleneck: encoding high-dimensional data (e.g., images) quickly requires too many qubits, and density-matrix operations scale as $4^N$ with the qubit number $N$. This limits expressive representations under practical simulation budgets. Inspired by latent diffusion, we mitigate this issue by learning in a compact latent space that can be embedded with only a small number of qubits.

In this work, we propose a scalable hybrid pipeline that brings mixed-state quantum diffusion into classical generation by combining (i) a classical autoencoder (AE) for dimensionality reduction \cite{hinton2006reducing} and (ii) a mixed-state quantum diffusion model in the learned latent space \cite{kwun2025mixed}. 
Concretely, a classical AE compresses high-dimensional samples into low-dimensional latent codes, which can be embedded into an $N$-qubit Hilbert space. 
The quantum diffusion model then learns the generative distribution over these latent quantum states, and samples are decoded back to pixel space by the AE decoder. 
This ''latent-space'' strategy echoes the motivation of classical latent diffusion models \cite{rombach2022high}, while leveraging quantum-state diffusion dynamics and mixed-state modeling.

A key algorithmic contribution is a simplified reverse update strategy: instead of learning an explicit predictor for $\rho_{t-1}$, we predict $\rho_0$ at time $t$ and obtain $\rho_{t-1}$ by a one-step backward propagation via \eqref{eq: reverse t-1}, rather than learning an explicit predictor for $\rho_{t-1}$. 
This mirrors the practical advantage of $x_0$-parameterizations in classical diffusion, but is implemented here at the density-matrix level. 
Intuitively, predicting $\rho_0$ provides a consistent global target (the ''clean'' state) across timesteps and allows the reverse update to explicitly incorporate the known structure of the forward channel through the analytic backward step, thereby reducing the learning burden of timestep-specific predictors.

\textbf{Contributions.} Our main contributions are:
\begin{enumerate}
    \item \textbf{Hybrid latent quantum diffusion for scalable generation:} we introduce an AE--QDM pipeline that mitigates the qubit bottleneck by learning a compact latent representation suitable for quantum-state encoding.
    \item \textbf{Mixed-state quantum diffusion backbone:} we build on the MSQuDDPM framework \cite{kwun2025mixed} to model forward noising as quantum channels and reverse denoising as a learnable quantum process, enabling generative learning over latent density operators.
    \item \textbf{Reverse update via $\rho_0$ prediction:} we implement the reverse step by predicting $\rho_0$ at time $t$ and analytically propagating one step backward using \eqref{eq: reverse t-1}, rather than training an explicit $\rho_{t-1}$ predictor.
    \item \textbf{Classical generative demonstrations:} we apply the proposed quantum diffusion model to classical image generation on MNIST \cite{lecun2002gradient}, demonstrating an end-to-end hybrid generative pipeline.
\end{enumerate}

The remainder of this paper is organized as follows: Section \ref{sec:preliminary} reviews the necessary quantum foundations. 
Section~\ref{sec:QDM} introduces the mixed-state quantum diffusion framework, describes the AE-based latent embedding, and outlines the classical generation pipeline. 
Experimental evaluations are provided in Section \ref{sec:Experiments}. 
Finally, Section \ref{sec:conclusions} concludes the paper. 

\begin{figure*}
    \centering
    \includegraphics[width=\linewidth]{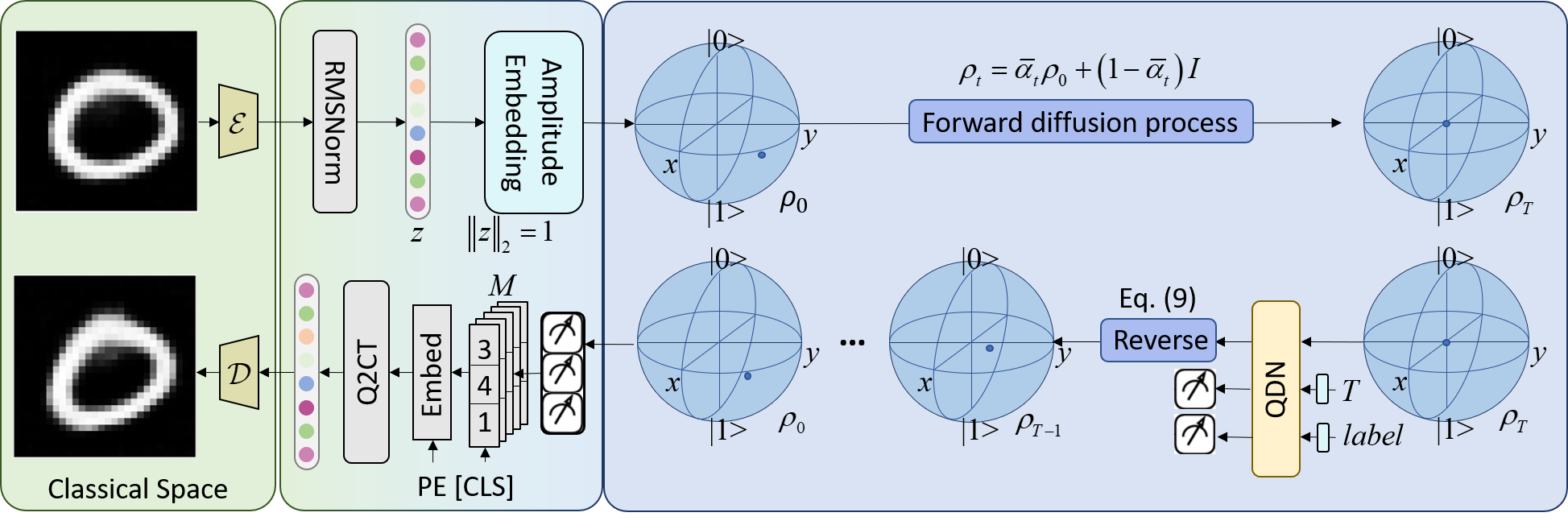}
    \caption{Overall framework: the forward process is defined in Eq.~\eqref{eq: forward}, and the corresponding conditional reverse update is given in Eq.~\eqref{eq: reverse t-1}. The quantum denoising network is described in Sec.~\ref{subsec:reverse}, and the classical pipeline is presented in Sec.~\ref{sec:classical part}.}
    \label{fig:frame}
\end{figure*}

\section{Preliminary}\label{sec:preliminary}

\subsection{Maximum Entropy State and Depolarizing Channel}\label{subsec:depolarizing}

In information theory, entropy quantifies the uncertainty of a system. For a quantum system, the entropy is usually analyzed by the von Neumann entropy:
\begin{align}
    S(\rho) = -\operatorname{tr}(\rho \ln \rho). 
    \label{eq:von Neumann entropy}
\end{align}
Among all density matrices $\rho$, the one that maximizes the von Neumann entropy is called the maximum entropy state (i.e., the completely mixed state) $\rho_{\text{mix}} \triangleq \frac{I}{d}$,
which represents the most conservative estimate of a system's state given partial information. It corresponds to the completely unbiased, least informative quantum state. 

In practice, quantum systems inevitably interact with their environment, which degrades coherence and introduces quantum noise. The depolarizing channel is a standard noise model that captures uniform random perturbations acting on a qubit during storage or transmission. 

For a $N$-qubit quantum system with density matrix $\rho$, the depolarizing channel $\mathcal{E}$ with depolarization probability $p$ is defined as:
\begin{align}
    \mathcal{E}(\rho) = (1-p) \rho + p \rho_{\text{mix}}.
    \label{eq:depo}
\end{align}

\subsection{Quantum Measurement: The Pauli-6 POVM}
An important example of positive operator-valued measures (POVMs) in quantum science is the Pauli-6 POVM, which is implemented as follows: (i) Randomly choose one of the three Pauli observables with equal probability $\tfrac{1}{3}$; (ii) Perform a projective measurement of the selected observable. 

The Pauli-6 POVM is informationally complete for a single qubit, meaning that the measurement statistics $\{p(m)\}$ contain sufficient information to fully reconstruct the quantum state $\rho$. This comes from the fact that any single-qubit density matrix can be expressed in the Bloch representation:
\begin{align}
    \rho = \frac{1}{2}\left(I + r_x X + r_y Y + r_z Z\right), 
    \label{eq: bloch rep}
\end{align}
where $(r_x, r_y, r_z)$ is the Bloch vector and can be calculated by the expectation of Pauli-6 POVM outputs. Thus, by repeating the measurement and estimating the outcome probabilities, one can determine the complete quantum state.

For an $N$-qubit system, one can apply the Pauli-basis randomization locally: the measurement setting is a string $s\in\{X,Y,Z\}^N$, and the outcome is a string $b\in\{\pm1\}^N$. Similar to single-qubit case, the local random Pauli measurement scheme with $s_i\in\{X,Y,Z\}$ is informationally complete.

In our work, we adopt the Pauli-6 scheme. The resulting collection $(s,b)$ provides a classical representation of the quantum state that is rich enough for our downstream tasks.

\section{Quantum Diffusion Model}\label{sec:QDM}

The overall framework of our model is illustrated in Fig.~\ref{fig:frame}. In the rest of this section, we describe each component in detail. 

\subsection{Forward Process}\label{subsec:forward}

As discussed in the previous section, depolarizing channel describes a natural process of adding noise to the quantum system. Thus, we consider letting original state $\rho$ go through a depolarizing channel at each timestep $t$ \cite{kwun2025mixed}: 
\begin{align}
    \rho_t = \alpha_t \rho_{t-1} + (1 - \alpha_t) \rho_{\text{mix}}, 
    \label{eq: forward}
\end{align}
where $\alpha_t\in[0,1]$ is the step-dependent noise strength. This forward process gradually "dilutes" the initial state $\rho_0$ toward the maximum entropy state $\rho_{\text{mix}}$. 

Analogous to the classical denoising diffusion probabilistic models, our forward process results in a direct mapping from the initial state $\rho_0$ to the state $\rho_t$ at any step $t$ in a closed form: 
\begin{align}
    \rho_t = \bar{\alpha}_t \rho_0 + (1 - \bar{\alpha}_t) \rho_{\text{mix}}, \quad \bar{\alpha}_t = \prod_{s=1}^t \alpha_s.
\end{align}

\subsection{Training loss}

Although the quantum relative entropy (QRE) is more closely related to the Kullback-Leibler (KL) divergence used in classical diffusion models, QRE cannot be defined on pure states, which are exactly what we want to generate through the quantum diffusion model for our task. Thus, we use the other widely used similarity measure—fidelity. For two density matrices $\rho$ and $\sigma$, the fidelity is defined as
\begin{align}
F(\rho, \sigma) = \left( \operatorname{tr} \sqrt{ \sqrt{\rho} , \sigma \sqrt{\rho} } \right)^2 .
\label{eq: def of fidelity}
\end{align}
The fidelity takes values in $[0, 1]$, attaining $1$ if and only if $\rho = \sigma$, and reaching zero for orthogonal states. It provides a direct and physically meaningful measure of "closeness" between quantum states. Therefore, we adopt a loss function based on fidelity between the generated state and the target state:
\begin{align}
\mathcal{L}_{\text{QDDPM}} = \mathbb{E}_{\rho_0} \left[\sum_{t=0}^{T-1} -\log F(\rho_{0|t}, \rho_{0})\right] ,
\label{eq: fidelity loss}
\end{align}
where $\rho_{0|t}$ indicates the denoised quantum state at timestep $t$. Here, we use the negative logarithm form so that the loss approaches zero as the fidelity approaches one, aligning with the minimization objective of the classical diffusion loss. 

\subsection{Reverse Process}\label{subsec:reverse}

Given the mixing state $\rho_t$ at timestep $t$, prior quantum diffusion works usually learn a separate model to directly predict previous state $\rho_{t-1}$. 
In contrast, in classical denoising diffusion (e.g., DDPM), predicting the clean image is found to be a better choice
 \cite{li2025jit}. 

\begin{figure*}[ht]
  \centering
  \begin{subfigure}{0.45\textwidth}
    \centering
    \includegraphics[width=\linewidth]{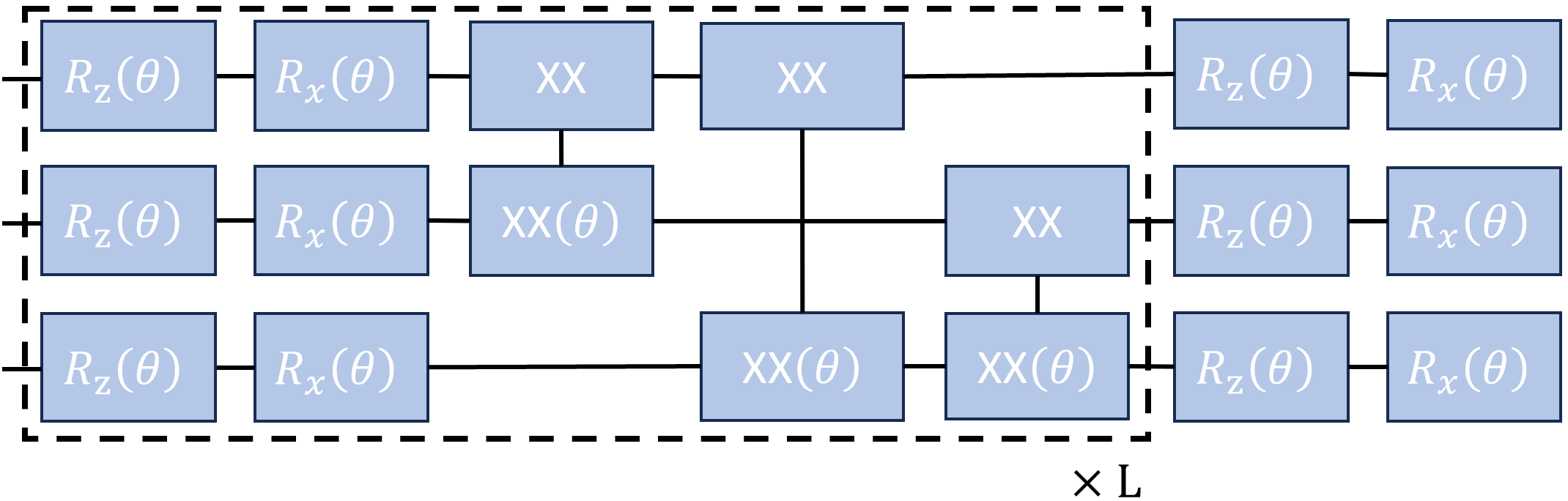}
    \caption{}
    \label{fig:denoise}
  \end{subfigure}\hfill
  \begin{subfigure}{0.45\textwidth}
    \centering
    \includegraphics[width=\linewidth]{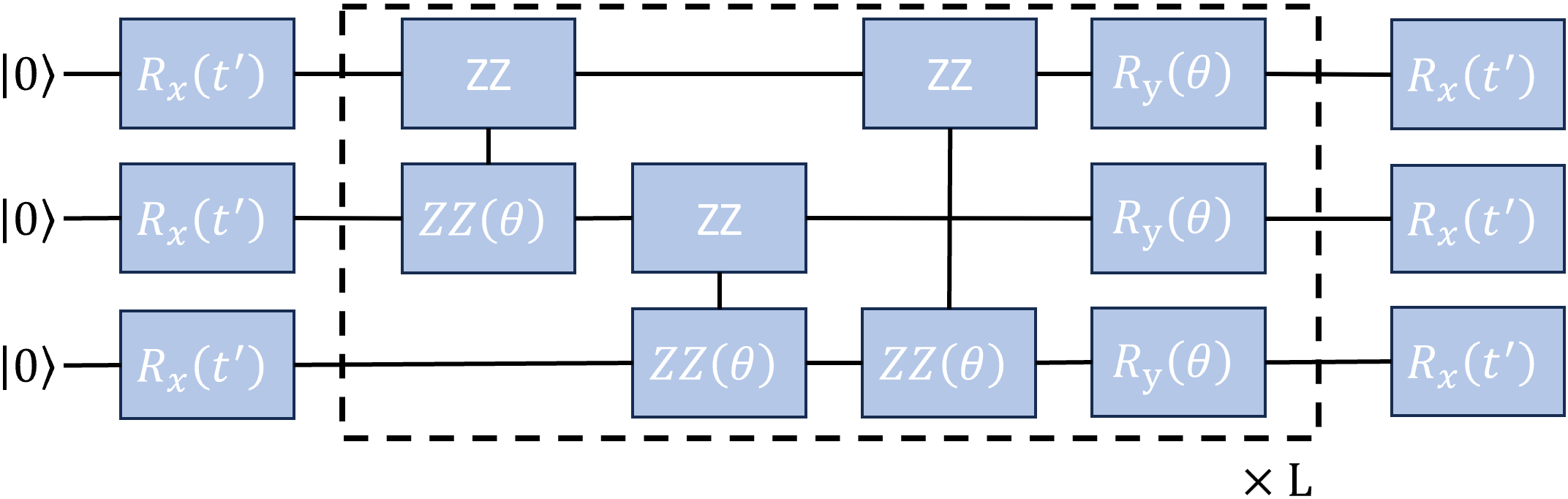}
    \caption{}
    \label{fig:embedding}
  \end{subfigure}
  \caption{Circuit architectures used in our model ($N=3$). (a) Denoising network using $R_z,R_x$ and fully connected $XX$ gates. (b) Embedding network using $R_x,R_y$ and fully connected $ZZ$ gates; $t'\in[0,\pi]$ denotes the normalized time/label input. In both circuits, $\theta$ denotes learnable parameters and $L$ is the number of layers.}
  \label{fig:circuits}
\end{figure*}

Motivated by this principle, we adopt the same strategy in the quantum setting: at each timestep $t$, our network outputs an estimate of the initial density matrix  $\rho_{0|t}$. 
Concretely, the construction of our quantum denoising network shown in Fig. \ref{fig:denoise}. It takes the current prediction $\rho_t$ together with auxiliary qubits: a time-embedding qubit and a label-embedding qubit (if conditional generation is required); see Fig. \ref{fig:embedding} for structure of the embedding circuits. After which the two auxiliary qubits are measured to produce $\rho_{0|t}$. Notably, the stochasticity at each reverse step is provided intrinsically by quantum measurement, rather than by injecting external noise.

We then compute $\rho_{t-1}$ from $\rho_t$ and $\rho_{0|t}$ using the following conditioned formula: 
\begin{align}
    \rho_{t-1}
    = \frac{1-\bar{\alpha}_{t-1}}{1-\bar{\alpha}_t}\,\rho_t
    + \frac{(1-\alpha_t)\bar{\alpha}_{t-1}}{1-\bar{\alpha}_t}\,\rho_{0|t}.
    \label{eq: reverse t-1}
\end{align}
Therefore, the reverse update can be implemented by predicting $\rho_0$ at time $t$ and propagating one step backward via \eqref{eq: reverse t-1}, rather than learning an explicit predictor for $\rho_{t-1}$.

\subsection{Converting Quantum States to Classical Distributions via POVM}

Although our quantum diffusion model operates entirely within the quantum state space, the final generated output must often be utilized or interpreted in a classical context. To bridge this gap, we employ the Pauli-6 POVM to perform a final measurement that maps the generated quantum state to a classical probability distribution. 

Additionally, in theoretical analysis or simulations where the full density matrix of the state is accessible, one can directly extract the quantum state's vector representation. Since our quantum diffusion model begins with pure states (as the initial quantum states are generated via eigenvector embeddings), and the reverse diffusion process is specifically designed to recover these pure states, the output density matrix $\rho_{\text{final}}$ is expected to be (approximately) pure, which means it has one eigenvalue close to 1 while all other eigenvalues are near 0. By performing an eigendecomposition and taking the principal eigenvector, we can recover a pure state vector $|\psi_{\text{final}}\rangle$ that fully encapsulates the quantum information: 
\begin{align}
|\psi_{\text{final}}\rangle = \arg\max_{|\psi\rangle} \langle \psi | \rho_{\text{final}} | \psi \rangle.
\end{align}
This vector can then be mapped to a classical feature vector (e.g., the $2^N$-dimensional complex amplitude vector) for further analysis or use. 

\subsection{Classical Neural Network}
\label{sec:classical part}
\subsubsection{Autoencoder}
We utilize a classical autoencoder to reduce the dimensionality of input data, such as images, before passing it to the quantum diffusion model. The autoencoder compresses the image into a compact latent space, which is then mapped into a small-qubit Hilbert space suitable for quantum processing. The encoder consists of three convolutional layers, progressively downsampling the input, followed by a fully connected network that produces the latent code $z$, capturing the essential features of the image. Finally, we normalize the output feature to meet the requirements of amplitude embedding.

The decoder mirrors the encoder, using transposed convolutions to reconstruct the original image from the latent code. The final output is passed through a sigmoid activation to ensure that the pixel values are between 0 and 1. This architecture allows for efficient encoding and decoding of images, enabling the quantum model to work within practical qubit limits, thus making the approach scalable for generative tasks in real-world settings.

\subsubsection{Quantum-to-Classical Transformer}
The Quantum-to-Classical Transformer (Q2CT) aims to reconstruct image features from quantum measurements by leveraging a transformer-based architecture \cite{tang2025quadim}. The process begins by embedding the quantum measurements, which are represented as integer values, into a continuous vector space using an embedding layer. Next, positional encodings are added to the embedded measurements to distinguish between different qubits, capturing the sequential relationships inherent in the quantum data.

To facilitate the reconstruction of the classical image features, a special [CLS] token is introduced \cite{dosovitskiy2021an}. This token serves to aggregate the information from all the quantum measurements, allowing the transformer to process and blend them effectively. The output from the transformer is then used to reconstruct the classical feature of the image, leveraging the learned relationships between the quantum measurements and their corresponding classical features. This approach enables efficient quantum-to-classical feature reconstruction, bridging the gap between quantum measurement and classical data representation for decoding.

\section{Experiments}
\label{sec:Experiments}
\subsection{Dataset and Experimental Setup}
To evalute the effectiveness of our proposed model, we conduct experiments on the MNIST dataset. MNIST remains a fundamental benchmark in the field of generative modeling due to its well-defined structure and clear evaluation metrics. The dataset consists of 70,000 grayscale images of handwritten digits (0-9), divided into a training set of 60.0000 examples and a test set of 10,000. Each sample is a $28\times 28$ pixel image, normalizaed to a fixed size where the digit is centered in the image. While visually simple, MNIST provides a rigorous testbed for assessing a model’s ability to capture discrete class distributions and produce sharp, high-contrast topological structures. In our study, it serves as the primary baseline for comparing direct $\rho_0$ prediction against iterative refinement strategies.

Our architecture utilizes an autoencoder with convolutional hidden dimensions of 32, 64, and 128, complemented by linear blocks of 256, 256, and 8. The autoencoder is trained using the Adam optimizer for 100 epochs with a learning rate of $10^{-3}$ and a batch size of 2000. For the QDN, we set the time embedding repetition $L_{\tau}=5$ and circuit repetition $L_{U}=1$, with total diffusion timesteps $T=30$. The QDN employs 3 qubits for feature embedding and 3 qubits for timestep embedding; in label-conditioned scenarios, this is adjusted to 2 qubits for timesteps and 1 for labels. The QDN and the Q2CT (configured with a 128-dimensional latent space, 4 attention heads, and 4 layers) are optimized via Adam with learning rates of $10^{-1}$ and $4 \times 10^{-4}$, respectively.

We employ a multi-faceted evaluation strategy comprising FID, KID, and IS scores. These metrics collectively capture the trade-off between visual quality and sample variety, where minimized FID/KID and maximized IS represent optimal model performance.

\subsection{Results}
\begin{figure}
    \centering
    \includegraphics[width=0.8\linewidth]{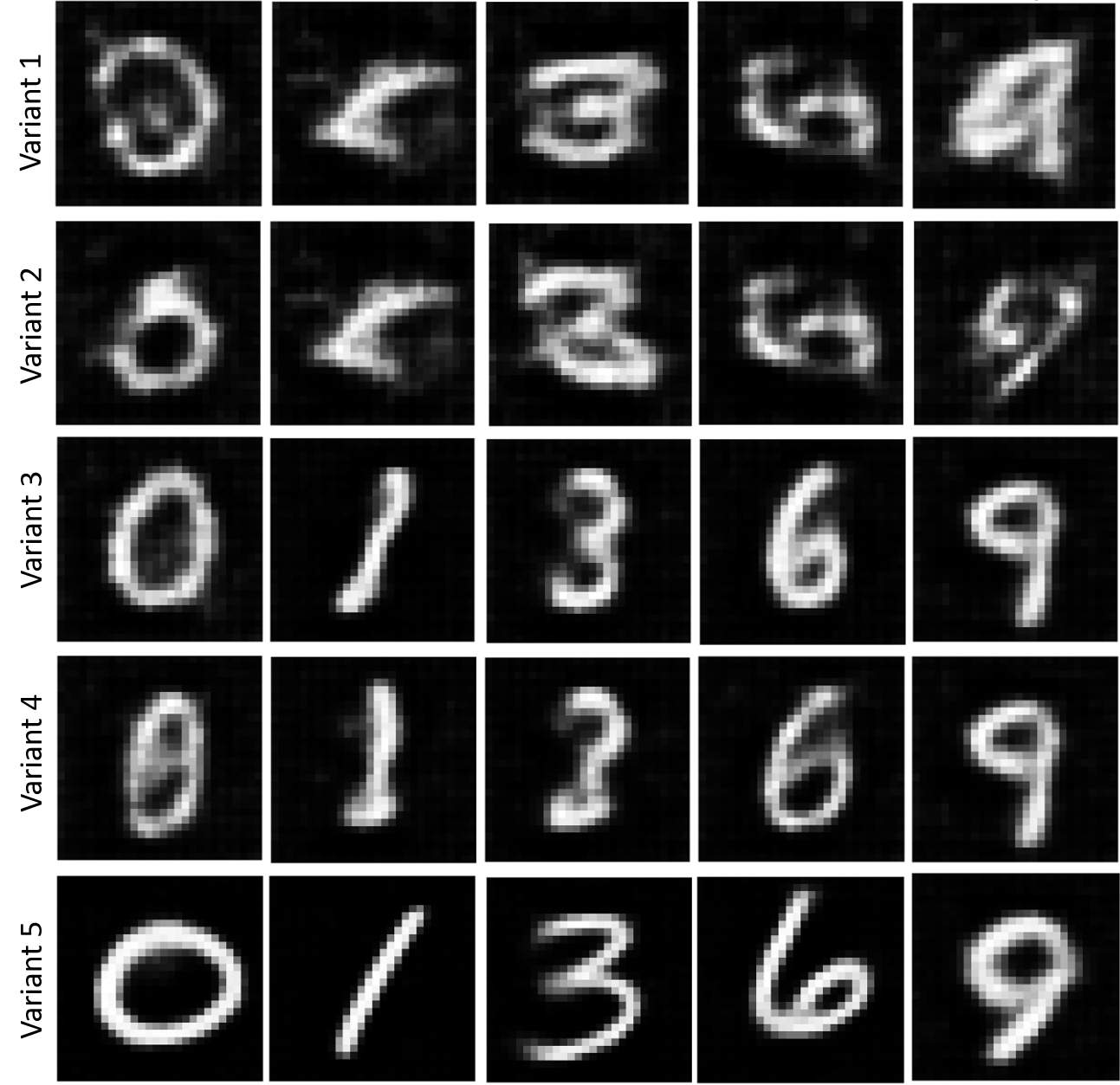}
    \caption{Illustrations about the generated images by different variants.}
    \label{fig:generated fig}
\end{figure}

We first evaluate our model’s generative capacity using both quantitative metrics and visual quality. The first two rows of Fig.~\ref{fig:generated fig} show representative samples under different configurations. A key contribution is our study of the reverse-process prediction target: we compare our proposed strategy of directly predicting the clean state $\rho_0$ with the standard step-by-step approach that predicts $\rho_{t-1}$. 

As shown in Table \ref{table:comparison}, the direct $\rho_{0}$ prediction (Variant 1) demonstrates superior efficiency and competitive accuracy compared to the iterative $\rho_{t-1}$ method (Variant 2), achieving better KID ($0.311$) and IS ($1.70$) scores. This suggests that our model effectively learns the global manifold mapping in a single step rather than relying on incremental refinements.

However, as observed in the visual results in Figure \ref{fig:generated fig}, images generated solely by the QDN exhibit noticeable noise and artifacts. This is likely due to the limited denoising capacity of the quantum circuit when handling complex distributions. In contrast, incorporating the Q2CT significantly enhances both quantitative and qualitative performance. As evidenced by Variants 3 and 4, the addition of the transformer-based architecture reduces the FID from $298$ to $247$ and further improves the KID and IS scores. This indicates that the Q2CT effectively captures long-range dependencies and refines the latent representations produced by the QDN.

To further explore the controllability of our framework, we investigate label-conditioned generation (Variant 5). While the FID ($252$) remains slightly higher than the best unconditional model, the qualitative samples in the bottom row of Figure \ref{fig:generated fig} reveal a distinct improvement in visual clarity.

Interestingly, the conditioned samples appear noticeably sharper than those produced by the unconditional model and even surpass the sharpness of the original MNIST images. This phenomenon suggests that the conditioning signal serves as a strong structural prior, not only guiding class identity but also exerting a potent denoising effect that forces the synthesis of idealized, high-contrast digit structures.



\begin{table}[tbh]
    \centering
    \caption{Generation performance of different variants under different settings.}
    \setlength{\tabcolsep}{5pt}
    \vskip 0.1in
    \begin{tabular}{@{}c|ccc|ccc@{}}
    \hline
       Variant &  Predict & Q2CT & Label & FID ($\downarrow$) & KID ($\downarrow$) & IS ($\uparrow$)  \\
    \hline
       1 & $\rho_0$ & & & 298 & \textbf{0.311} & \textbf{1.70}  \\
       2 & $\rho_{t-1}$ & & & \textbf{296} & 0.322 & 1.49  \\
       \hline
       3 & $\rho_0$ & \checkmark & & \textbf{247} & \textbf{0.267} & \textbf{1.84} \\
       4 & $\rho_{t-1}$ & \checkmark & & 251 & 0.274 & 1.69 \\
       \hline
       5 & $\rho_0$ & \checkmark & \checkmark & 252 & 0.291 & 1.80 \\
    \hline
        
    \end{tabular}
    \label{table:comparison}
\end{table}


\section{Conclusions}
\label{sec:conclusions}
This work advances quantum diffusion toward practical classical image generation by applying a physically grounded quantum diffusion framework to real image synthesis. We also improve the reverse process over prior formulations: rather than predicting $\rho_{t-1}$ step by step, we directly predict the clean state $\rho_0$ at each timestep and compute the reverse update via the forward-induced posterior. Experiments show this strategy consistently improves generation quality, validating the proposed design. 
Despite the small-qubit regime, our results show that the proposed QDM can still learn meaningful sample distributions and generate structured outputs, and that label conditioning further improves controllability and sample quality. These findings suggest that quantum diffusion is a viable and promising direction for generative modeling, and that its potential will become increasingly practical as quantum hardware and simulation tools continue to scale.

\vfill\pagebreak
\bibliographystyle{IEEEbib}
\bibliography{strings,refs}

\end{document}